\documentclass[]{article}

\usepackage{times}

\usepackage[english]{babel}
\usepackage[utf8x]{inputenc}
\usepackage[T1]{fontenc}
\usepackage{enumitem}
\usepackage[margin=0.75in]{geometry}

\usepackage{natbib}
\usepackage{amsmath}
\usepackage{graphicx}
\usepackage{subcaption}
\usepackage{dsfont}

\usepackage{float}
\usepackage{graphicx}
\usepackage{epstopdf, epsfig}
\usepackage{flushend}
\usepackage[dvipsnames]{xcolor}
\usepackage{amsmath}
\usepackage{latexsym}
\usepackage{caption}
\usepackage{color,soul}
\usepackage{physics}
\usepackage[normalem]{ulem}

 \usepackage{physics}
 \usepackage{amssymb}
 \usepackage{amsmath}

\graphicspath{{figures/}}
\usepackage[colorlinks=true, allcolors=blue]{hyperref}
\usepackage{xcolor}
\usepackage{pgfplots}
\usepackage{bm}
\usepackage{todonotes}
\usepackage{comment}

\usepackage{algorithm,amsmath,algpseudocode,algcompatible,setspace}

\definecolor{light-yellow}{rgb}{1,1,0.75}

\title{Information FOMO: The unhealthy fear of missing out on information. A method for removing misleading data for healthier models}

\author{Ethan Pickering, Themistoklis P. Sapsis\\
Massachusetts Institute of Technology\\
Department of Mechanical Engineering} 

\begin{document}

\maketitle

\begin{abstract}

Misleading or unnecessary data can have out-sized impacts on the health or accuracy of Machine Learning (ML) models. We present a Bayesian sequential selection method, akin to Bayesian experimental design, that identifies critically important information within a dataset, while ignoring data that is either misleading or brings unnecessary complexity to the surrogate model of choice. Our method improves sample-wise error convergence and eliminates instances where more data leads to worse performance and instabilities of the surrogate model, often termed sample-wise ``double descent''. We find these instabilities are a result of the complexity of the underlying map and linked to extreme events and heavy tails. 

Our approach has two key features. First, the selection algorithm dynamically couples the chosen model and data. Data is chosen based on its merits towards improving the \textit{selected} model, rather than being compared strictly against other data. Second, a natural convergence of the method removes the need for dividing the data into training, testing, and validation sets. Instead, the selection metric inherently assesses testing and validation error through global statistics of the model. This ensures that key information is never wasted in testing or validation. The method is applied using both Gaussian process regression and deep neural network surrogate models.


\end{abstract}

\section{Introduction}

What we term, ``Information FOMO'', or Fear Of Missing Out, is the unhealthy tendency to use all available data for fear of missing out on information. However, more data is not always better. Data may present misleading information, or even worse, lead to overfitting and, as we show here, induce instabilities in the surrogate map. These cases may be linked to the concept of sample-wise ``deep double descent''\cite{opper1995statistical,belkin2019reconciling, nakkiran2021deep}, where more data does not result in better models. However, this phenomena can more generally be attributed to slow sample-wise convergence.

\begin{figure}
\vspace{0.5cm}
\centering
\includegraphics[width=0.5\linewidth,trim={0.25cm 0cm 0cm 0cm},clip]{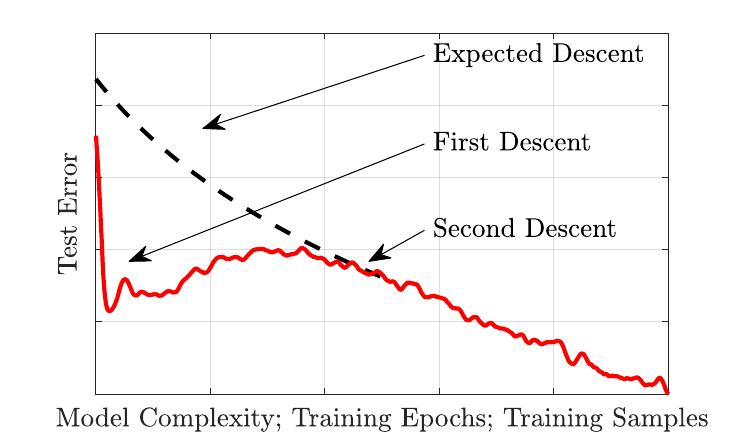}
\caption{\textbf{Double Descent does not follow the expected descent of modern ML techniques.} Modern ML expects test error to decrease with model complexity, training epochs, and training samples, yet, in practice, the descent is not monotonic.}
\label{fig:DD_Cartoon}
\end{figure}

The general concept of double descent, figure \ref{fig:DD_Cartoon}, refers to test errors that first undergo a descent, then an \textit{ascent} in error, followed by a second and final descent in error. This phenomena is observed with respect to model complexity (i.e. number of layers or layer width), training epochs, or training samples \cite{nakkiran2019more}, and has been both theoretically appreciated and empirically observed in numerous studies \cite{opper1995statistical,spigler2018jamming,geiger2019jamming, belkin2019reconciling,  advani2020high, nakkiran2021deep, hastie2022surprises, pickering2022discovering}. Although all three manifestations of double descent can lead to significant errors, the first two, model complexity and training epochs, can be avoided through straight forward model parameter studies, such as ``early stopping'' strategies \cite{yao2007early} in either training time \cite{heckel2020early} or model size. Training sample size, however, is nearly always fixed and often too small for sufficient cross-validation studies. As such, double descent with small and fixed samples presents a substantial threat to ML techniques in real-world applications.


Despite this threat to ML techniques, sample-wise double descent in deep models has received much less attention than model complexity and training time. This is likely due to the limitation that fixed datasets are not large enough to perform validation studies. The majority of literature surrounding sample-wise double descent is recognized in ridgeless regression 
\cite{hastie2022surprises,nakkiran2020optimal} or random features regression \cite{mei2022generalization}. In ridgeless  regression, \cite{nakkiran2020optimal}, double descent is directly related to an instability of the solution to an ill-posed and noisy dataset when dimensions $d$ are equal to fitting parameters $n$. This notion of an instability of the surrogate model is precisely what we find here.   

Instabilities may be addressed in many cases as improved generalization error due to sample-wise double descent has been observed through appropriate $\ell_2$ regularization of the model in ridge regression \cite{hastie2022surprises,nakkiran2020optimal} or data-dependent regularizers in deep models \cite{wei2019data,wei2019improved}, however, the latter provides modest improvements of a few percent for most presented cases. Our work here follows in the vein of data-dependent regularizers, where the data itself is seen as a component to the level of complexity of the model, to eliminate instabilities of the surrogate model and improve error convergence.

While not applied to concepts of sample-wise double decent, there are several approaches for selecting subsets of training data to \textit{reduce training cost} on large datasets that draw parallels to our approach. To reduce neural network training cost, random uniform sampling \cite{shalev2011pegasos,shalev2013stochastic}, importance sampling \cite{allen2016even,nesterov2017efficiency,needell2014stochastic,katharopoulos2018not,zhao2015stochastic}, and adaptive importance sampling \cite{csiba2015stochastic, perekrestenko2017faster,alain2015variance} are used to choose only a fraction of the data for training evaluations. Adaptive importance sampling is closest to our approach, using probability distributions to define sample importance and update the training set at each iteration \cite{stich2017safe}. While quite effective, these approaches are designed for efficiently updating and training the network weights, not for neglecting unnecessarily complex data or improving the stability of a general surrogate model. Our application considers not only neural networks, but also Gaussian processes, which do not undergo stochastic gradient descent for training. 

In Section \ref{sec:Alg} we propose a method, inspired by Bayesian experimental design, for eliminating double descent through identifying, and ignoring, data that brings unnecessary complexity to the model. This is done by iteratively selecting a subset of data for improving the model by scoring the dataset with respect to its likely information gain and predictive uncertainty. This idea allows us to ignore data whose predictive uncertainty and likely information gain are small. We find that doing so brings substantial improvements to the convergence of test mean-squared error (MSE), as well as the often unappreciated log-PDF error, which emphasizes accurate prediction of extreme events.

The method also converges without the need for testing or validation data while only using a fraction of the data. As the method's selection criteria ignores much of the data, convergence in the training set directly implies convergence in test error. The approach also demonstrates that only a small subset of the data, 1 out of every 20 samples, provide useful information. Together, the unnecessary need for splitting data intro training, testing, and validation sets and the minimal selection of data for optimal error properties suggest this approach may unlock the use of many small and fixed datasets throughout various applications.

These findings are demonstrated first in Section \ref{sec:res_GP} on a simple, nonlinear $1D$ function, with a tunable extreme event parameter, approximated by Gaussian process (GP) regression that presents a visual example of sample-wise double descent. To our knowledge, sample-wise double descent has yet to be observed for GP regression and our tuning from non-extreme to extreme behavior underscores how data can induce instabilities in the surrogate map. We then extend this approach in Section \ref{sec:res_DNN} to a larger and significantly more complex problem with 20,000 training samples and Deep Neural Networks (DNNs) taking the role of surrogate model (specifically, DeepONet \cite{lu2021learning}). Section \ref{sec:discussion} concludes our study discussing the generality and implications of the approach to limited and sparse datasets as well as any Bayesian or ensemble surrogate model.

\vspace{0.5cm}
\section{Methods}
\vspace{0.5cm}
\subsection{FOMO algorithm}
\label{sec:Alg}
The ``FOMO'' sequential selection method for fixed datasets is outlined in algorithm~\ref{alg:1}. The method is initialized by first choosing the $n_a$ ``optimal'' samples of the dataset. These are determined by calculating the likelihood ratio, $w(\mathbf{x}) = p_{\mathbf{x}}(\mathbf{x}) / p_\mu(\mu)$, or information metric, on all samples after approximating the output PDF, $p_\mu$, through one surrogate model (e.g. DNN or GP) training on the complete input-output dataset. Once the top scoring information metric samples are chosen, the surrogate model is trained with these $n_a$ samples and emit predictions of $\mu$ and $\sigma^2$ for all observed data pairs. The surrogate model is also leveraged to construct the approximate output PDF (the output PDF is approximated by $10^7$ Latin Hypercube samples (LHS) whose output is predicted by the surrogate model). The product of the information metric and predictive variance provide the acquisition values, $a(\mathbf{x})$ \cite{blanchard2021output}, and the best $n_a$ samples, amongst the whole dataset, are augmented to the training set. Finally, because \textit{all} samples are given acquisition scores, duplicates, i.e. a current training sample is chosen again, are removed. The process is then repeated until the algorithm ceases to acquire any unique samples.

\begin{algorithm}[t]
  \caption{Sequential selection algorithm for fixed data.}\label{algorithm1}
  \begin{algorithmic}[1]
    \STATE \textbf{Initialize:} Train surrogate model on \textbf{full} dataset of I-O pairs 
    \STATE Build weight function $w(\mathbf{x})$, choose $n_{\mathrm{init}}$ best samples from the dataset.
    \STATE Retrain surrogate model on only $n_{\mathrm{init}}$ best samples.

    \STATE \textbf{for} $n=1$ \textbf{to} $n_{iter}$ \textbf{do}
    \STATE \hspace{0.25cm} From \textbf{full} dataset, select best point $\mathbf{x}_{n_{a}}$  for acquisition function $a(\mathbf{x})$
    \STATE \hspace{0.25cm} Augment training dataset and \textbf{remove} duplicates.
    \STATE \hspace{0.25cm} Retrain surrogate model on augmented dataset.
    \STATE \textbf{end for}
    \STATE \textbf{return} Final surrogate model \textit{and} training dataset.
  \end{algorithmic}
  \label{alg:1}
\end{algorithm}

\vspace{0.5cm}

\section{Results}
\vspace{0.5cm}

We show that the FOMO approach improves error convergence in both mean-square error (MSE) and log-PDF error and eliminates model instabilities for GP and DNN surrogate models in both low and high dimensional problems, without the need for test or validation data.

\subsection{$1D$ Example with Gaussian Process Regression} 
\label{sec:res_GP}

\begin{figure}[t!]
$a)$ \hspace{5.25cm} $b)$ \hspace{5.25cm} $c)$ \phantom{PPPPPPPP} \\
\includegraphics[width=1\linewidth]{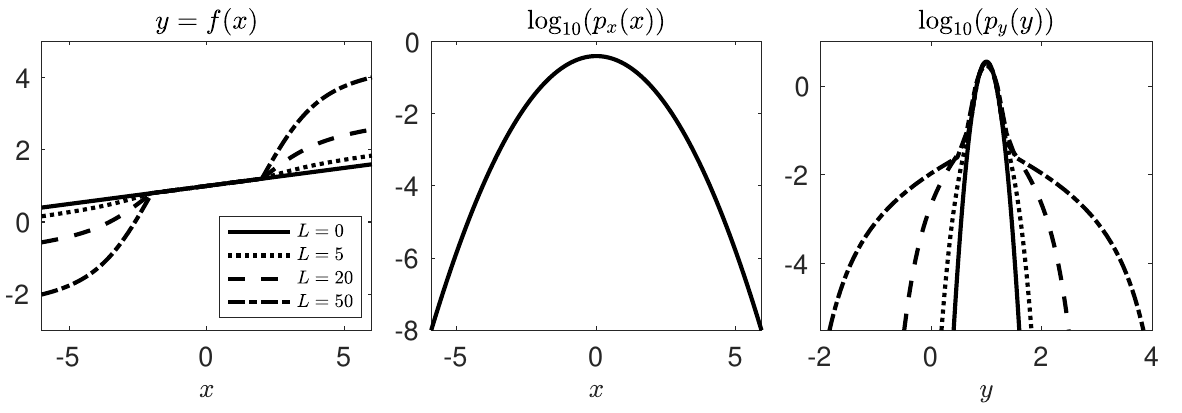}
\caption{$a)$ the true nonlinear solutions $y=f(x)$, with respect to random variable $x$ for nonlinear coefficients of $L=0,5,20,50$, $b)$ the Gaussian PDF of $x$, and $c)$ the non-Gaussian PDF of response variable $y$ with heavy tails for each nonlinear case.}
\label{fig:GP_True}
\end{figure}

As with nearly all ML tasks, we wish to predict a state, $y$, from an observed input, $x$, by approximating the underlying map, $y = f(x)$. Approximating this map with a surrogate model may require substantial data depending on the complexity and dimension of the input space (e.g. $x$ is multi-dimensional vector of inputs, $\mathbf{x}$). Figure \ref{fig:GP_True}$~a)$ presents test functions we wish to approximate: a piece-wise nonlinear function of varying degrees (see \S~\ref{sec:N1D}) with a linear core and nonlinear edges to emulate rare dynamical instabilities initiated at high magnitudes (varied by nonlinear coefficient, $L=0,5,20,50$). The input variable, $x$, is a Gaussian random variable with mean 0 and standard deviation of 1, whose probability distribution function (PDF), $p_x$, is shown in figure \ref{fig:GP_True}~$b)$. We may then calculate the PDF of $y$ via standard weighted Gaussian kernel density estimators (KDE) as,
\begin{equation}
    p_{f}(y) = \mathrm{KDE}(\mathrm{data}=y, \mathrm{weights}=p_x(x)), 
\end{equation}
which displays a Gaussian core and heavy tails in figure \ref{fig:GP_True}~$c)$, indicating rare and extreme events.



\begin{figure}[t]
\centering
\begin{minipage}{.475\textwidth}
$a)$ \hspace{4cm} $b)$ \phantom{P}  \\
\includegraphics[width=0.95\linewidth,trim={0cm 13.1cm 0cm 0cm},clip]{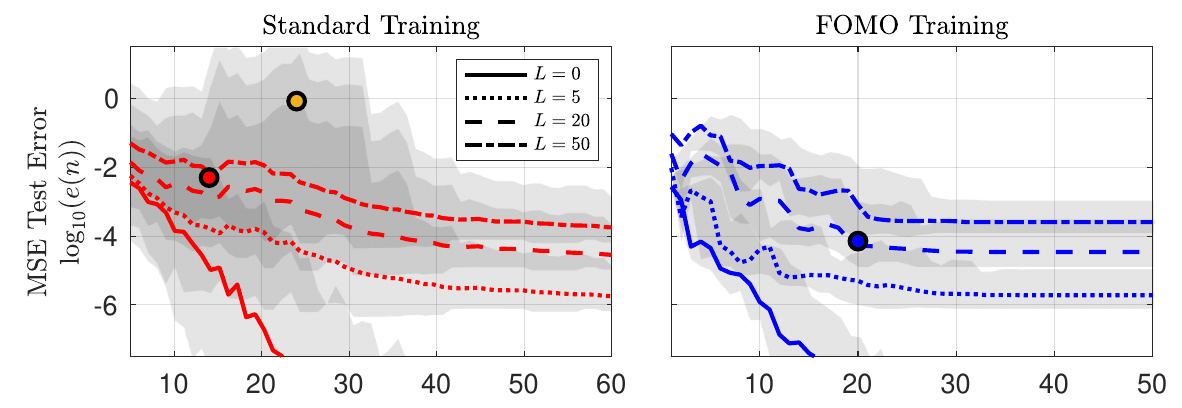} \\
$c)$ \hspace{4cm} $d)$ \phantom{P} \\
\includegraphics[width=0.95\linewidth,trim={0cm 7.1cm 0cm 6.9cm},clip]{figs/GP_Metric_Plots.pdf} \\
$e)$ \hspace{4cm} $f)$ \phantom{P}
\includegraphics[width=0.95\linewidth,trim={0cm 0.2cm 0cm 13cm},clip]{figs/GP_Metric_Plots.pdf}
\caption{\textbf{GP FOMO model improves error convergence, is superior to early stopping, and converges without testing or validation data.} The mean normalized MSE and log-PDF errors (min and max values shaded) of 100 experiments with  randomly chosen data samples ( $a)$ and $c)$ ), and with sequential selections ( $b)$ and $d)$ ) over four nonlinear coefficients $L=0,5,20,50$. $e)$ provides a comparison of the approximated and true solution for the errors denoted in $a)-d)$ and $f)$ is the number of chosen data samples by iteration for 100 independent sequential searches from $b)$ and $d)$.}
\label{fig:FOMO_GP}
\end{minipage}
\hspace{0.5cm}
\begin{minipage}{.475\textwidth}
\vspace{-0.25cm}
$a)$ \hspace{4cm} $b)$ \phantom{P}  \\
\includegraphics[width=0.95\linewidth,trim={0cm 13.1cm 0cm 0cm},clip]{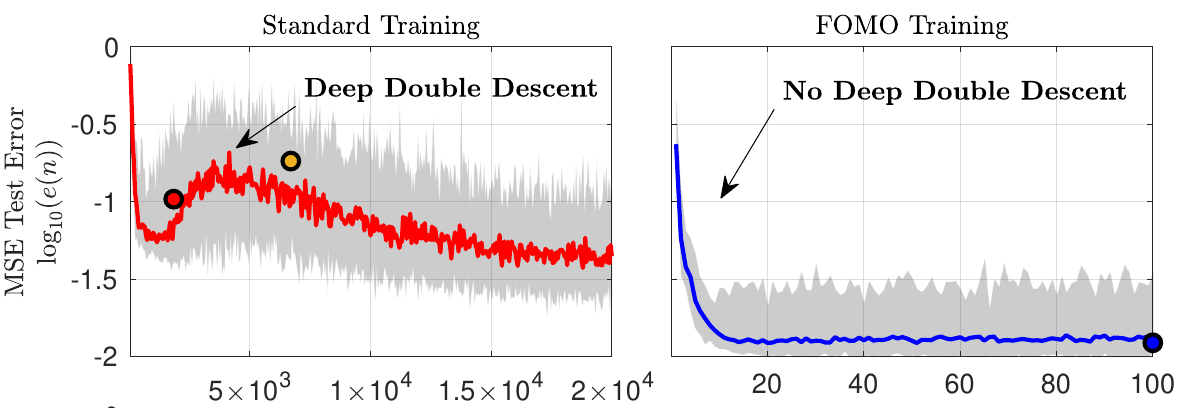} \\
$c)$ \hspace{4cm} $d)$ \phantom{PPPPPP} \\
\includegraphics[width=0.95\linewidth,trim={0cm 7.1cm 0cm 6.9cm},clip]{figs/LHS_Metric_PDF_Samples_Alternate.pdf} \\
$e)$ \hspace{4cm} $f)$ \phantom{PPPPPP} \\
\includegraphics[width=0.95\linewidth,trim={0cm 0.2cm 0cm 13cm},clip]{figs/LHS_Metric_PDF_Samples_Alternate.pdf}
\caption{\textbf{DNN FOMO model eliminates double descent, is superior to early stopping, and converges without testing or validation data.} The mean normalized MSE and log-PDF errors (with min and max values shaded) of 25 experiments with randomly chosen data points on a Latin Hypercube ( $a)$ and $c)$ ), and with sequential selections ( $b)$ and $d)$). $e)$ provides a comparison of the predicted and true output distributions for three errors denoted in $a)-d)$ and $f)$ the number of chosen data samples by iteration for 25 independent sequential searches from $b)$ and $d)$.}
\label{fig:FOMO}
\end{minipage}
\end{figure}



In figure \ref{fig:FOMO_GP}, we test and observe training errors for 100 independent experiments for our two training approaches approximating \ref{fig:GP_True}$~a)$ with a GP surrogate model, detailed in \S~\ref{sec:GP_method}, given $n$ samples (sampled from a uniform distribution), ranging from 5-60, of observed input and output data, $\mathcal{D} =  \{x_i, y_i \}_{i=1}^{n}$. Figures \ref{fig:FOMO_GP}~$a)$ and \ref{fig:FOMO_GP}~$c)$ present the normalized MSE and log-PDF test errors (see \S~\ref{sec:error_calc} for error calculation details), respectively, against the number of samples in standard training, while \ref{fig:FOMO_GP}~$b)$ and \ref{fig:FOMO_GP}~$d)$ report the errors versus iteration using the ``FOMO'' algorithm, described in \S~\ref{sec:Alg}. In each figure the median error is highlighted with shaded regions indicating the minimum and maximum errors. 

Poor error convergence, i.e. sample-wise double descent, is observed for both metrics under standard training using GPs for the simple $1D$ problem for nonzero nonlinear coefficients, while the FOMO approach clearly improves convergence. For this $1D$ case, we will forego detailed early-stopping/training/testing/validation implications for the following DNN example and focus on the $L=20$ 1D representations of sample-wise double descent and the instabilities experienced by the surrogate model. We highlight 3 points in plots $a)-d)$. The red, gold, and blue points present representative examples of early stopping point, double descent, and FOMO training, respectively, from the \textit{same} training set. Figure \ref{fig:FOMO_GP}$~e)$ provides the mean solution, $\mu(x)$, for each and their associated training samples. While the red, $n_x = 14$, solution is relatively accurate, the addition of 10 more samples leads to an unstable solution at $n_x = 24$. The FOMO approach avoids this instability and presents a substantially superior model solution by sample 10, though $n_x = 24$ is shown to compare against the double descent solution. This observation of model instabilities through standard training are also observed for the other two nonlinear coefficients, $L=50$ and $L=5$, with more and less severity, respectively. These coefficients lead to heavier tails and stronger extreme events, while $L=0$ emits no extreme events and already possess fast error convergence with standard training.

Figure \ref{fig:FOMO_GP}$~f)$, presents another key feature of the FOMO algorithm, choosing only a fraction of the possible samples. The algorithm only adds samples that bring stable improvements to the model and ignores those that do not. We expand on this idea in the higher-dimensional case using DNNs next.


\subsection{Dispersive Nonlinear Wave Model with Deep Neural Networks}
\label{sec:res_DNN}

We now move to a more complex case that seeks to learn a dispersive nonlinear wave model that has the form of a one-dimensional partial differential equation originally proposed by Majda, McLaughlin, and Tabak (MMT) \cite{majda1997one} for the study of 1D wave turbulence. The initial conditions are chosen randomly from an $8D$ subspace: $u(x_{L},t) = \mathbf{x} \mathbf{\Phi}(x_L)$, where $x_L$ refers to the spatial variable, while $x$ or $\mathbf{x}$ represents the random coefficients (an $8D$ vector), which are assumed to follow a known distribution. Our aim is to identify a maximum future wave height $|Re(u(x,t=\tau))|_{\infty}$ (see \S~\ref{sec:NLS} and \S~\ref{sec:map_definition} for details on MMT and the wave height map). While GPs could be used as a surrogate, a companion study \cite{pickering2022discovering} showed DNNs are superior for the complexity of this problem and are used here, see \S~\ref{sec:DNN_method} for DNN details. Additionally, this example requires many training samples, at least at an order of magnitude of 1,000 or more, and are computationally intractable for standard off-the-shelf GP applications compared to off-the-shelf DNNs. 

Similar to the previous example, figure \ref{fig:FOMO} presents the training errors of 25 independent experiment of both training approaches for approximating the underlying map with a DNN given $n$ samples (sampled via Latin Hypercube sampling, LHS), ranging from 50-20,000, of observed input and output data, $\mathcal{D} =  \{ \boldsymbol{u}_i(x_{L}), y_i \}_{i=1}^{n}$. Figures \ref{fig:FOMO}~$a)$ and \ref{fig:FOMO}~$c)$ present the normalized MSE and log-PDF test errors, respectively, against the number of samples, while \ref{fig:FOMO}~$b)$ and \ref{fig:FOMO}~$d)$ report the errors versus iteration using the ``FOMO'' algorithm. In figures \ref{fig:FOMO}~$a)$ and \ref{fig:FOMO}~$c)$, early stopping only considers 1500 samples and a similarly accurate error is not observed again until an order of magnitude higher, at 15,000 samples. Even at the early stopping error, figure $e)$ shows a drastic under prediction of the most probable states. More concerning, if an unconservative splitting of the 20,000 samples is taken, i.e. 6,700 samples for training, testing, and validation, then full training leads to errors near the peak of double descent. In addition to the early stopping errors, the peak of double decent also leads to extreme over predictions of high magnitude events, shown in figure \ref{fig:FOMO}~$e)$. 

The ``Information FOMO'' results shown in figures \ref{fig:FOMO}~$b)$ and \ref{fig:FOMO}~$d)$, and found via algorithm \ref{alg:1}, do not suffer from these difficulties. Instead, the FOMO approach provides the most accurate model (see figure~\ref{fig:FOMO}~$e)$), does not undergo sample-wise double descent, converges quickly, and only uses  1/20th of the data to achieve these results. We stress the latter observation as it is a direct result of algorithm \ref{alg:1}, which, at each iteration, only selects data that specifically eliminates uncertainty in the model or provides essential information. Consequently, data that does not bear beneficial information to the model is deemed unnecessary and ignored.

Figure \ref{fig:FOMO} ~$f)$ demonstrates the ability of the algorithm to ignore data and converge to a set of optimal training data that, critically, emits converged error metrics in both MSE and log-PDF error. At each iteration, the algorithm finds a user-specified $n_a$ samples (here $n_a = 50$) that provide the greatest information gain amongst the \textit{entire} dataset. This means that even those samples that are amongst the training set are considered. As more samples are added to the training set, the algorithm finds that the remaining samples provide \textit{less} information than \textit{known} samples, permitting an ability to disregard those remaining samples. The convergence of the training set results in a convergence of information, which ultimately leads to a convergence in the test error (figures~$b)$ and $d)$). Critically, this implies that the FOMO approach does not require data be split into training, testing, and validation groups. Rather, all pertinent information may be extracted from the entire dataset for the most accurate model.

\begin{figure}
\centering
\includegraphics[width=0.75\linewidth]{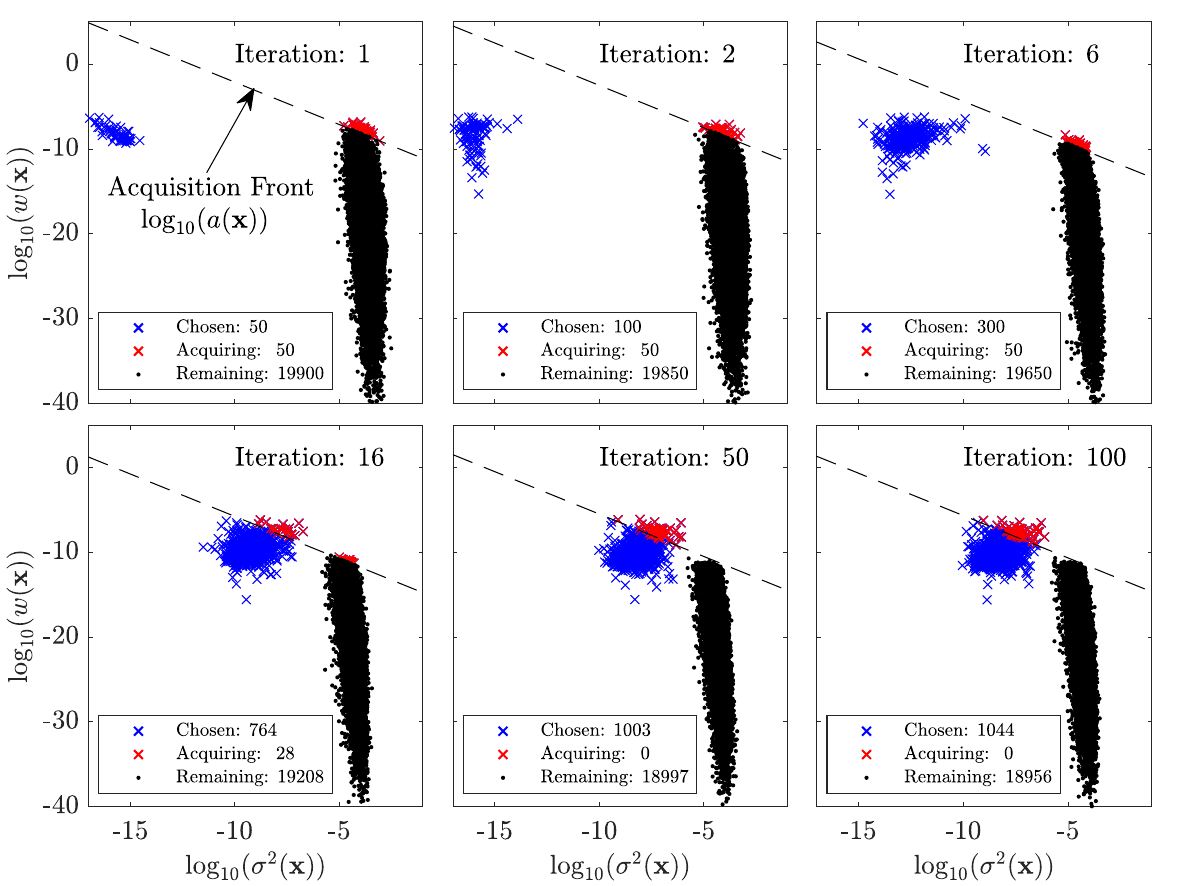}
\caption{\textbf{Important and misleading/unnecessary datasets show clear separation.} A representative example of the iterative selection process, where data is sequentially acquired. The acquisition front indicates the acquisition score of the 50th optimal point at each iteration. Those above are acquired and those already chosen remain in the training set.}
\label{fig:AcuisitionFront}
\end{figure}

\textbf{Optimal data selection through a dynamic partitioning of samples as informative or unnecessary.}
Figure \ref{fig:AcuisitionFront} demonstrates how algorithm \ref{alg:1} selects the most informative samples, while ignoring the rest. The $y-$axis denotes our metric for information, the likelihood ratio \cite{blanchard2021output},
\begin{equation}
    w(\mathbf{x}) = \frac{p_\mathbf{x}(\mathbf{x})}{p_\mu(\mu)},
\end{equation}
while the $x-$axis presents the predictive variance, $\sigma^2(\mathbf{x})$, found amongst the ensemble of independently weight-initialized DNN predictions. The product of these two quantities is the acquisition function proposed by \cite{blanchard2021output,sapsis2020output}, $a(\mathbf{x}) = w(\mathbf{x})\sigma^2(\mathbf{x})$. As shown in \cite{sapsis2022optimal}, the adopted acquisition function balances information with uncertainty that guarantees optimal convergence in the context of Gaussian Process Regression. Here it is employed as a measure to optimally acquire samples for training the DNN. In effect, this approach allows us to measure model risk versus model capacity \cite{angluin1983inductive,blumer1987occam,vapnik1999nature,hastie2009elements,belkin2019reconciling}.

Walking through the iterations of figure \ref{fig:AcuisitionFront} we can see how the approach dynamically partitions the data into informative and unnecessary. Starting at iteration 1, only 50 samples, denoted in blue are used to train the model (See algorithm \ref{alg:1} for how these samples are chosen). Unsurprisingly, such little data allows each DNN to train these samples to machine precision and gives a predictive uncertainty amongst the training samples of $\log_{10}(\sigma_{Chosen}^2(\mathbf{x})) \approx -16$, while the predictions of the remaining samples are 12 orders of magnitude larger, as expected, and vary widely in the likelihood ratio metric. We can then acquire the largest $n_a$ samples, with respect to the acquisition function, retrain the model with the addition of the new samples, and repeat the process. Here we chose $n_a=50$ and the 50th point sets what we define as the ``acquisition front'', denoting a line of constant acquisition value between samples the model wishes to acquire and those it does not. Iteration 2 and 6 show that as more data is acquired into the chosen/training set the uncertainty in the chosen set increases, while the uncertainty in the remaining set decreases. The latter observation is simply due to the increased training data that reduces generalization error and thus uncertainty. The former observation, although expected, has significant implications for the FOMO method. As the training set increases, the degree of over-parameterization of each DNN reduces and discrepancies between the DNNs training errors increase. This behavior is critical to identifying when the DNN parameterization becomes stressed and susceptible to instabilities related to slow error convergence or double descent. This is shown in the following iterations.

Iterations 16, 50, and 100, demonstrate the ability of the model to ignore data when DNN parameterization has reached a critical threshold. At iteration 16 the uncertainty of the chosen data has substantially increased, while the information contained in the remaining data has decreased, leading to an acquisition front the intersects both the chosen and remaining sets. For this iteration, only 28 of the 50 samples are novel, while 22 samples already exist in the chosen data. This means that the model begins to recognize that it already contains the most pertinent information \textit{and} its uncertainty in this data is approaching an unacceptable level. By iteration 50 the model has begun to entirely ignore the remaining data, as the model is sufficiently uncertain about its own, more informative, training data and does not see a reason to add further stress into the model. Over the next 50 iterations, the stochastic nature of the DNN training permits 43 new samples to join the chosen dataset. However, considering the average of acquisition is less than one for the 50 potential acquisition samples at each iteration, the algorithm is easily converged by iteration 50.

\begin{figure}
\centering
$a)$ \hspace{5cm} $b)$ \phantom{PPPPPPPPPPPPPPPPPP} \\
\includegraphics[width=0.8\linewidth]{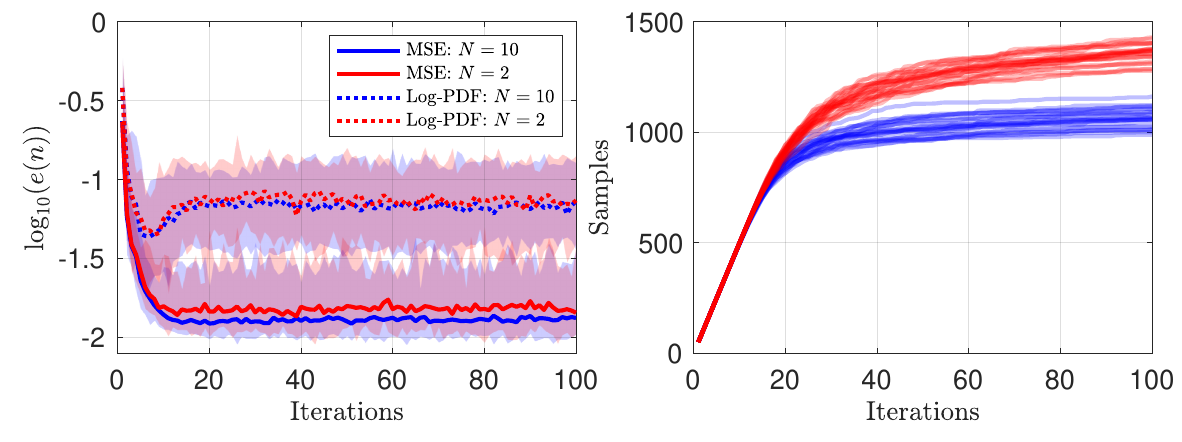}
\caption{\textbf{Shallow ensembles perform well.} $a)$ The median MSE and log-PDF errors (with min and max values shaded) of 25 experiments using DNN ensemble sizes of $N=10$ and $N=2$. $b)$ the number of chosen data samples by iteration for 25 independent sequential searches for both $N=10$ and $N=2$. }
\label{fig:N_Effect}
\end{figure}

\textbf{Shallow ensembles are cheap and perform well.}
An ensemble of DNNs, differed only in their weight initialization \cite{lakshminarayanan2016simple}, are employed to calculate the predictive uncertainty. Although the validity of such approaches is hotly debated, \cite{wilson2020bayesian} and also \cite{pickering2022structure}, have argued that DNN ensembles provide a very good approximation of the posterior. While the previous results have all been for an ensemble of $N=10$ DNNs, figure \ref{fig:N_Effect} shows that shallow ensembles of only $N=2$ perform nearly as well at 1/5 the computational cost. Figure \ref{fig:N_Effect}~$a)$ shows that the median and range of both the MSE and log-PDF errors are nearly identical, with small advantages going to the larger ensemble. The greatest difference between the two is the number of acquired samples in figure~\ref{fig:N_Effect}~$b)$, where $N=2$ chooses approximately 25\% more samples. The surprising ability of shallow ensembles of $N=2$ to perform well in active learning schemes was also observed in \cite{pickering2022discovering}.

\section{Discussion}
\label{sec:discussion}

\textbf{Slow sample-wise error convergence, or double descent, a traditionally unavoidable phenomenon for fixed datasets, is eliminated in both the MSE and log-PDF errors for both GP and DNN surrogate models.} This Bayesian-inspired approach, ignores data that brings unwarranted instability to the chosen model that otherwise results in an increase in test/generalization error with increased sample size, an instability which is often unrecoverable. As most datasets in real-world applications are fixed, removing sample-wise double descent is critically important. This method not only substantially reduces the errors in normalized MSE from a median of 20\%, to 1\%, but also the log-PDF error by 4-fold. An often unappreciated metric, the log-PDF error appropriately weights high-magnitude, rare phenomena that ML methods struggle to accurately predict \cite{sapsis2021statistics,pickering2022discovering}.

\textbf{The approach is model-agnostic and statistically driven, requiring a dynamic interaction between the model and the data.} The statistical metrics used to select data means that regardless of the chosen surrogate model (e.g. DNN, Gaussian process regression, etc.) or parameters (e.g. layers, neurons, training epochs, activation functions, kernels, etc.), the model and the dataset have an opportunity to ``discuss'' the inherent deficiencies of the model and the short comings of the data. As only statistics are used to guide the iterative data selection process, the approach only requires that the surrogate model of choice emit a mean and variance prediction. Although rigorous predictions of the quantities, such as Gaussian process regression, are clear surrogate model candidates, our work here shows that simple, ensemble DNN provide sufficient predictions for success.

\textbf{Shallow DNN ensembles are simple in implementation, scalable, and computationally tractable.} Ensemble DNNs are simple to implement. They only require that DNN weights are randomly initialized, the default setting in all DNN architectures. Despite this simple approach, we find that shallow ensembles of DNNs, even just two, provide sufficient predictive uncertainties to eliminate double descent. DNNs also provide ideal scaling in data size and inputs parameters, unlike robust methods such as Gaussian processes. Finally, although the iterative process requires many new DNNs to be trained, the DNNs are trained on only a small subset of the data, 1/20th here, during the selection process. Together, ease of implementation, modest scaling, shallow ensembles, and reduced training data are compelling for real-world application.


\textbf{Convergence of the approximated output PDF ensures convergence of training samples and removes the need for testing and validation sets.} The surrogate model output PDF, $p_\mu(\mu)$ directly, and dynamically, informs the information metric (i.e. likelihood ratio) for acquiring new samples. Once the surrogate output PDF converges, the information metric does as well. Consequently, the acquisition front becomes set, only the predictive variance provides an avenue for acquiring more samples, and the remaining samples above this threshold are acquired, ending the selection. While it is traditional to consider the point-wise MSE in validation and testing, consisting of 2/3 of the data (i.e. 12,500 here), the surrogate output PDF is approximated through $10^6$, or more, test samples. The surrogate output PDF provides a global measure of the model generalization error, rather than a sparse and limited pointwise comparison. Thus, we argue that the cost of losing information to a local comparison tool, by setting aside 2/3 or more of available data, compared to a global metric is a far greater and unnecessary risk. 


\textbf{An optimal sparse representation of the underlying map is preferred to a plentiful, but arbitrary representation, opening the door to numerous datasets with various deficiencies, whether that be from unnecessary or limited data.} Although the likelihood ratio is likely appropriate for most physical systems, other choices may be optimal for different applications. Further work is also necessary to extend the method's performance in pathological cases and sufficiently noisy data.

\textbf{Sample-wise double descent in the nonlinear context appears to have a fundamentally different meaning compared to traditional discussions of double descent.} While we do not provide any theoretical findings here, the observation of sample-wise double descent for Gaussian Process regression does not follow current reasoning.
Double descent is typically associated with a crossover from under- to over-parameterization of DNNs. However, a GP in $1D$ has only two tunable parameters (three if noise is considered), meaning the under- to over-parameterization occurs over 2 samples, yet double descent is observed from 5-30 samples. From our observations of both GPs and DNNs, we believe double descent is due to the underlying map and the distribution of sample data. If the map is sufficiently complex, i.e. nonlinear, and data is not appropriately distributed in the input space, as is the case when randomly partitioned, than the surrogate model may possess instabilities. Of course, in the $1D$ case, it is trivial to propose a uniform distribution, but in higher-dimensional cases, such as the MMT case here, a sufficient uniform distribution of samples is infeasible. The FOMO method circumvents this by finding the distribution of fixed samples that best recover the global, output PDF, which includes both the Gaussian core and heavy tails where nonlinearities lie.

\textbf{Instabilities of the surrogate model induced by the underlying map complexity.} Specifically, the complexity refers to the degree to which extremes and heavy tailed statistics are present in the underlying map. By varying a tunable nonlinear coefficient, we show that sample double descent is not present for a non-extreme Gaussian map while double descent becomes increasingly apparent for maps with extremes and heavy tails. As more extremes are present, the surrogate models become more susceptible to instabilities. Using our approach, these data-induced instabilities are avoided. 

\textbf{Drawbacks to this approach.} While the FOMO approach improves the ability to eliminate double descent, double descent may still be unavoidable without enough samples. We list a few limitations below:
\begin{enumerate}
    \item If the surrogate solution for the entire dataset is located at the peak of double descent, the initial output PDF will emit a suboptimal FOMO initialization. Several approaches could help mitigate this: initialize with the early stopping solution, propose a physics-based assumption of the underlying output PDF, or propose further iterative FOMO procedures for initial data refinement.
    \item The above underscores that although the approach will extract the most relevant samples, if no subset of the samples are sufficient to characterize the map, FOMO will not improve performance.
    \item While the GP case provides a compelling $1D$ representation of double descent, the instability could be removed by introducing a nonzero noise term. Based on our knowledge that the problem is noiseless, this would be an incorrect application of noise, but, it would remove the instability.
    \item Other regularization approaches for instabilities could be employed for DNNs, such as physical constraints or physics-based loss functions (e.g. PINNs \cite{raissi2019physics}), but their efficacy for faster error convergence or sample-wise double descent is unknown. Additionally, the use of regularization approaches does not invalidate use of the FOMO method.
    \item The method assumes a known input distribution $p_\mathbf{x}$. This distribution could be approximated or rationalized from the input data. If incorrect, the performance would suffer. Future work is necessary to determine the sensitivity of the method with respect to errors in the input distribution.
\end{enumerate}

In summary, we believe the FOMO approach is compelling for many practical problems. It mitigates sample-wise error convergence in both MSE and log-PDF, eliminates instabilities introduced by data, removes traditional train/test/validation by converging towards global statistics, and is simple and scalable via flexible off-the-shelf DNNs and shallow ensembles. Future work should continue to explore this performance in additionally challenging problems and on noisy data. While we generally discuss small datasets here, the FOMO approach could also be implemented to compress data and information for intractably large datasets. This a tremendous challenge for domains such as climate where individual climate simulations can be on the order of terabytes to petabytes.

\section{Appendix}
\subsection{Map definition and data}
\label{sec:map_definition}

\subsubsection{Nonlinear $1D$ Function}
\label{sec:N1D}
The piece-wise $1D$ equation is given by a linear function with logistic functions at each end,
\begin{align}
    y &= \bigg( ax + \frac{L}{1 + e^{-k(x+2)}} - L/2\bigg) / b +1  && x < -2 \\ 
    y &= ax/ b +1    && -2 \geq x \leq 2 \\
    y &= \bigg(ax + \frac{L}{1 + e^{-k(x-2)}} - L/2\bigg) / b +1  && x > 2,
\end{align}
where  $L = 20$, $k=1$, $a=1$, and $b=10$.

\subsubsection{Nonlinear Dispersive Wave Model}
\label{sec:NLS}

The underlying nonlinear system in our study is the Majda, McLaughlin, and Tabak (MMT) \cite{majda1997one} model, a dispersive nonlinear model that also includes selective dissipation for high wavenumbers, used for studying $1 \mathrm{D}$ wave turbulence. It is  described by
\begin{equation}
    i u_{t}=\left|\partial_{x}\right|^{\alpha} u+\lambda\left|\partial_{x}\right|^{-\beta / 4}\left(\left.\left.|| \partial_{x}\right|^{-\beta / 4} u\right|^{2}\left|\partial_{x}\right|^{-\beta / 4} u\right)+i D u,
    \label{eqn:MMT}
\end{equation}
where $u$ is a complex scalar, exponents $\alpha$ and $\beta$ are chosen model parameters, and $D$ is a selective Laplacian. Under appropriate choice of parameters its response is characterized by extreme events which occur intermittently. Specifically, the model contains a rich set of physical phenomena, from four-wave resonant interactions that produce both direct and inverse cascades \cite{majda1997one,cai1999spectral} and presents a unique utility as a physical model for extreme ocean waves, or rogue waves \cite{zakharov2001wave,zakharov2004one,pushkarev2013quasibreathers}.
We refer the reader to \cite{pickering2022discovering}, where identical parameters were used and to \cite{cousins2014quantification} for the numerical calculation technique. 

In defining an input-output problem, we are not interested in quantifying the entire MMT behavior, but a specific observation that does not possess any closed form solution. We define the input as  
\begin{equation}
    u(x,t=0) \approx \mathbf{x} \boldsymbol{\Phi}(x_L), \hspace{0.5cm} \forall \hspace{0.5cm} x_L \in [0,1)
\end{equation}
where $\mathbf{x} \in \mathbb{C}^m$ is a vector of complex coefficients and both the real and imaginary components of each coefficient are normally distributed with zero mean and diagonal covariance matrix $\Lambda$, and $ \{ \boldsymbol{\Lambda}, \boldsymbol{\Phi} (x)\} $ contains the first $m$ eigenpairs via a Karhunen-Loeve expansion of the correlation matrix defined by the complex and periodic kernel,
\begin{equation}
 k(x_L,x_L^\prime)   = \sigma_{u}^2 e^{i(x_L-x_L^\prime)} e^{-\frac{(x_L-x_L^\prime)^2}{\ell_u}},
\end{equation}
with $\sigma_u^2 = 1$ and $\ell_u = 0.35$. This gives the dimension of the parameter space as $2m$ due to the complex nature of the coefficients. For all cases, the random variable $\mathbf{x}_i$ is restricted to a domain ranging from -6 to 6, in that 6 standard deviations in each direction from the mean. Here we chose $m=4$, or 8 total independent and real variables. This value could easily be chosen to be higher, however, doing so requires an intractable amount of data to recover from deep double descent (see \cite{pickering2022discovering} for a $20D$ double descent example), further underscoring the threat sample-wise double descent brings to ML.

Finally, the output map is then defined as,
\begin{equation}
    f(\mathbf{x}) = || Re(u(x_L,t=T;\mathbf{x}))||_{\infty},
\end{equation}
where $T=20$. This map serves the purpose as a predictor for the largest incoming wave and, if trained appropriately, informs of a future dangerous rogue wave. 

The training data for each of the 25 independent experiments consist of 20,000 samples distributed throughout the $8D$ parameter space with Latin Hypercube sampling. Both the traditional training and FOMO algorithm share the same 25 independent datasets of 20,000 samples.

\subsection{Surrogate Models}

\subsubsection{Gaussian Process Regression}
\label{sec:GP_method}
Gaussian process regression \cite{rasmussen2003gaussian} is the ``gold standard’’ for Bayesian design. A Gaussian process, $\bar{f}(\mathbf{x})$, is completely specified by its mean function $m(\mathbf{\mathbf{x}})$ and covariance function $k\left(\mathbf{\mathbf{x}}, \mathbf{\mathbf{x}}^{\prime}\right)$. For a dataset $\mathcal{D}$ of input-output pairs ($\{\mathbf{\mathbf{X}}, \mathbf{y}\})$ and a Gaussian process with constant mean $m_{\boldsymbol{0}}$, the random process $\bar{f}(\mathbf{\mathbf{x}})$ conditioned on $\mathcal{D}$ follows a normal distribution with posterior mean and variance
\begin{equation}
\mu(\mathbf{\mathbf{x}})=m_{0}+k(\mathbf{\mathbf{x}}, \mathbf{\mathbf{X}}) \mathbf{K}^{-1}\left(\mathbf{y}-m_{0}\right)
\label{eqn:mean_GP}
\end{equation}
\begin{equation}
\sigma^{2}(\mathbf{\mathbf{x}})=k(\mathbf{\mathbf{x}}, \mathbf{\mathbf{x}})-k(\mathbf{\mathbf{x}}, \mathbf{\mathbf{X}}) \mathbf{K}^{-1} k(\mathbf{\mathbf{X}}, \mathbf{\mathbf{x}})
\label{eqn:cov_GP}
\end{equation}
respectively, where $\mathbf{K}= k (\mathbf{\mathbf{X}}, \mathbf{\mathbf{X}})+\sigma_{\epsilon}^{2} \mathbf{I}$. Equation \eqref{eqn:mean_GP} can predict the value of the surrogate model at any point $\mathbf{\mathbf{x}}$, and \eqref{eqn:cov_GP} to quantify uncertainty in prediction at that point \cite{rasmussen2003gaussian}. Here, we chose the radial-basis-function (RBF) kernel with automatic relevance determination (ARD),
\begin{equation}
k\left(\mathbf{\mathbf{x}}, \mathbf{\mathbf{x}}^{\prime}\right)=\sigma_{f}^{2} \exp \left[-\left(\mathbf{\mathbf{x}}-\mathbf{\mathbf{x}}^{\prime}\right)^{\top} \mathbf{L}^{-1}\left(\mathbf{\mathbf{x}}-\mathbf{\mathbf{x}}^{\prime}\right) / 2\right],
\end{equation}
where $\boldsymbol{L}$ is a diagonal matrix containing the lengthscales for each dimension and the GP hyperparameters appearing in the covariance function $(\sigma_{f}^{2}$ and $\boldsymbol{L}$ in \eqref{eqn:cov_GP} are trained by maximum likelihood estimation). Additionally, training the GP in equation \eqref{eqn:cov_GP} requires the inversion of the matrix $\mathbf{K}$. Typically performed by Cholesky decomposition, the inversion cost scales as $O(n^{3})$, with $n$ being the number of samples \cite{rasmussen2003gaussian,shahriari2015taking}. The prohibitive cost of this for large datasets (>$\mathcal{O}(10^3$)) is one reason DNNs are used for the MMT case.

\subsubsection{Ensemble Deep Neural Network}
\label{sec:DNN_method}

The DNN implemented here is a simple feed-forward neural network, built with \textsc{tensorflow} and using packages including \textsc{deepxde} and \textsc{deeponet} \cite{lu2021learning}. The network consists of 8 layers, 250 neurons, is trained for 1000 epochs, under a learning rate of $l_r=0.001$, ReLu activation functions, and a mean square error loss function. 

For each set of training data, an ensemble of $N$ Glorot normal randomly weight-initialized DNN models, each denoted as $\tilde{G}_{n}$, approximate the associated solution field $y$ for feature inputs $u$. This allows us to determine the point-wise variance of the models as
\begin{equation}
    \sigma^2(u) = \frac{1}{(N-1)} \sum_{n=1}^{N} (\tilde{G}_{n}(u) - \overline{\tilde{G}(u)})^2,
\end{equation}
where $\overline{\tilde{G}(u)}$ is the mean solution of the model ensemble. Finally, we must adjust the above representation to match the description for Bayesian design. In the case of traditional Bayesian design and GPs, the input parameters, $\mathbf{x}$, are random coefficients to a set of functions that represent a decomposition of a random function $u = \mathbf{x} \mathbf{\Phi}(x_{L}^1, ... x_{L}^m)$, where $x_{L}^1$ and $x_{L}^m$ represent the spatial boundaries of the function. Thus, the DNN description for UQ may be recast as:

\begin{align}
    &\sigma^2(\mathbf{x}) \nonumber  = \\ &\frac{1}{(N-1)} \sum_{n=1}^{N} \bigg(\tilde{G}_{n}(\mathbf{x} \mathbf{\Phi}(x_{L}^1, ... x_L^m)) - \overline{\tilde{G}(\mathbf{x} \mathbf{\Phi}(x_L^1, ... x_L^m))}\bigg)^2.
\end{align}

\subsection{Test error computation}
\label{sec:error_calc}
To compute the MSE and log-PDF test errors, we select two random sets of test samples, $10^2$ for the GP example and $10^5$ for the DNN example, $\mathbf{X}_{\mathrm{pdf}}$, pulled from the input PDF $p_\mathbf{x}$, and $\mathbf{X}_{\mathrm{lhs}}$, from Latin Hypercube sampling (random uniform for the GP case, $\mathbf{X}_{\mathrm{uni}}$), where $\mathbf{X}_{\mathrm{GP}} \in \mathbb{R}^{d \times 10^2}$ and $\mathbf{X}_{\mathrm{DNN}} \in \mathbb{R}^{d \times 10^5}$. The normalized MSE is computed using $\mathbf{X}_{\mathrm{pdf}}$ as input and $y_{\mathrm{pdf}} = f(\mathbf{X}_{\mathrm{pdf}})$ as output. The normalized MSE is then calculated as,
\begin{equation}
    e_{\mathrm{MSE}}(n) = \frac{ \sum_{i=1}^{n} (y_{\mathrm{pdf},i} - \mu_{\mathrm{pdf},i})^2}{\sum_{i=1}^{n} y_{\mathrm{pdf},i}^2} 
\end{equation}
where $n=10^5$. The log-PDF error uses the $y_{\mathrm{lhs}} = f(\mathbf{X}_{\mathrm{lhs}})$ input-output pairs and is calculated as
\begin{equation}
  e_{\mathrm{log-PDF}}(n) = \int | \log_{10} p_{\mu_n} (y) - \log_{10} p_f (y)| \text{d} y, 
\end{equation}
where both the true PDF, $p_f (y)$, and approximated PDF, $p_{\mu_n} (y)$, are found via a kernel density estimator as,
\begin{equation}
    p_{f}(y) = \mathrm{KDE}(\mathrm{data}=y_{\mathrm{lhs}}, \mathrm{weights}=p_\mathbf{x}(\mathbf{X}_{\mathrm{lhs}})), 
\end{equation}
and 
\begin{equation}
    p_{\mu}(y) = \mathrm{KDE}(\mathrm{data}=\mu_{\mathrm{lhs}}, \mathrm{weights}=p_\mathbf{x}(\mathbf{X}_{\mathrm{lhs}})).
\end{equation}
With the exception that the GP example uses the uniform sampling, $y_\mathrm{uni}, \mu_\mathrm{uni}, \mathbf{X}_\mathrm{uni}$.
We reiterate that the purpose of the log-PDF error is to insure that the approximation appropriately accounts for PDF tails, where rare, high magnitude events, i.e. extreme events, live.

\subsection*{Code and Data Availability}

Data and code pertaining to the FOMO sequential search algorithm will be made public upon publication. Currently, the code for the GP problem is provided and executable via this Google Colaboratory link: \href{https://colab.research.google.com/drive/1I74vzOPoGzaOmXpV6D2f9L3pTR2Brj6N?usp=sharing}{GP FOMO Google Colab}, while the DNN problem is provided through this Dropbox link: \href{https://www.dropbox.com/sh/60s9hsw6hqzpfsv/AABSgZdZm5Ujk0LQXno4_5eta?dl=0}{DNN FOMO Dropbox}.


\subsection*{Acknowledgments} The authors acknowledge support from DARPA (Grant No. HR00112290029) as well as from AFOSR (MURI grant FA9550-21-1-0058), awarded to MIT.

\bibliographystyle{abbrv}
\bibliography{references}

\appendix

\end{document}